\documentclass{article} % For LaTeX2e
\usepackage{iclr2026_conference,times}
\iclrfinalcopy % 展示作者信息，终稿形式
\usepackage{amsmath,amsfonts,bm}

\def\eqref#1{equation~\ref{#1}}
\def\1{\bm{1}}

\DeclareMathAlphabet{\mathsfit}{\encodingdefault}{\sfdefault}{m}{sl}
\SetMathAlphabet{\mathsfit}{bold}{\encodingdefault}{\sfdefault}{bx}{n}

\usepackage{wasysym}
\usepackage{hyperref}
\usepackage{url}
\usepackage{tcolorbox}
\usepackage{multirow,amssymb,amsmath,tabularx,graphicx} %zph add
\usepackage{booktabs} % Used to create more aesthetically pleasing table lines (\toprule, \midrule, \bottomrule)
\usepackage{colortbl}       
\definecolor{tablegray}{gray}{0.9} %0.9 indicates a 90% white mixture, which will be very light in color and suitable for use as a background
\usepackage{caption}
\usepackage{enumitem}
\usepackage{algorithm}
\usepackage{listings}
\usepackage{xcolor}

\usepackage{algpseudocode}
\usepackage{caption}

\title{PVPO: Pre-Estimated Value-Based Policy Optimization for Agentic Reasoning}

\author{Wenfeng Feng\ensuremath{^*}, Penghong Zhao\thanks{The first two authors contributed equally} , Guochao Jiang, Chuzhan Hao, Yuewei Zhang,\\
\textbf{Guohua Liu,} \textbf{Hao Wang}\protect \thanks{ Corresponding author} \\
Alibaba Cloud Computing \\
\texttt{\{wenfeng.fwf,zhaopenghong.zph\}@alibaba-inc.com}\\ \texttt {suoni@taobao.com,cashenry@126.com}
}
\begin{document}

\maketitle

\begin{abstract}
Critic-free reinforcement learning methods, particularly group policies, have attracted considerable attention for their efficiency in complex tasks. 
However, these methods rely heavily on multiple sampling and comparisons within the policy to estimate advantage, which may cause the policy to fall into local optimum and increase computational cost. 
To address these issues, we propose PVPO, an efficient reinforcement learning method enhanced by an  advantage reference anchor and data pre-sampling.
%%%%%
Specifically, we use the reference model to rollout in advance and employ the calculated reward score as a reference anchor.
Our approach effectively corrects the cumulative bias introduced by intra-group comparisons and significantly reduces reliance on the number of rollouts during training. 
Meanwhile, the reference model can assess sample difficulty during data pre-sampling, enabling effective selection of high-gain data to improve training efficiency.
Moreover, PVPO is orthogonal to other advanced critic-free RL algorithms, making it compatible with and complementary to these methods.
%%%%%
Experiments conducted on nine datasets across two domains demonstrate that PVPO achieves State-Of-The-Art (SOTA) performance. Our approach not only demonstrates robust generalization across multiple tasks, but also exhibits scalable performance across models of varying scales.
\end{abstract}

\section{Introduction}
Reinforcement Learning (RL) is a machine learning method for learning optimal policies through interaction with the environment. Policy optimization depends on accurately estimating the advantage function to improve the agent's actions.
In classic actor-critic frameworks, a critic network predicts state-value ($V$), which combines with action-value ($Q$) to compute the advantage and then guides policy updates. Recently, research has increasingly focused on more efficient critic-free architectures. These methods do not directly compute the absolute advantage. Instead, they build baselines for relative advantage, simplifying the training process and reducing resource consumption~\citep{shao2024deepseekmathpushinglimitsmathematical,feng2025groupingrouppolicyoptimizationllm}.

Grouping policies, as used in critic-free RL methods like GRPO~\citep{shao2024deepseekmathpushinglimitsmathematical}, become an important research topic. This is not only because they demonstrate superior performance, but also because the removal of the value model saves training resources, enabling researchers to train larger-scale models under limited hardware conditions. Although PPO and other actor-critic methods sometimes achieve higher accuracy, critic-free grouping policies are widely used for their practical efficiency. Some studies group by sample, running multiple trajectories within each group to compute relative advantage~\citep{zuo2025ttrltesttimereinforcementlearning, lyu2025hierarchicalbudgetpolicyoptimization}. Others group by action or timestep, enabling finer partitioning and more accurate baseline estimation~\citep{feng2025groupingrouppolicyoptimizationllm, li2025reporeplayenhancedpolicyoptimization}.
These methods can improve baseline accuracy for similar trajectories. However, grouping policies usually require more rollouts to boost performance, which greatly increases computational cost. 
% Methods like DAPO \citep{yu2025dapoopensourcellmreinforcement} try to reduce it by focusing on high-value data sampling, but mainly shift the resource use rather than actually lowering it. In the end, there is always a trade-off between training performance and computational cost.
Methods such as DAPO~\citep{yu2025dapoopensourcellmreinforcement} aim to mitigate this issue by prioritizing high-value data sampling. However, they primarily redistribute resource utilization rather than achieving a genuine reduction in overall resource consumption. We still need to achieve an effective trade-off between training performance and computational cost. To construct the relative advantage, some methods use state-independent baselines to generate advantage values for each action \citep{REINFORCE, ahmadian2024basicsrevisitingreinforcestyle}. GRPO ~\citep{shao2024deepseekmathpushinglimitsmathematical} and GiGPO~\citep{feng2025groupingrouppolicyoptimizationllm} compare the rewards of actions or trajectories within groups.
% In these approaches, the evaluation standard comes from the policy itself. As a result, policy optimization can get stuck in existing behavior patterns and fall into local optima.
In these approaches, the evaluation criterion is derived from the policy itself, which may cause policy optimization to become confined to existing behavior patterns and lead to local optima.
From a human learning perspective, rollout can be seen as repeated practice. Grouping policies resemble trial-and-error learning, where individuals often compare outcomes to a fixed \textbf{Reference Anchor} for more efficient learning. This anchor serves as an objective reference point, distinct from the idealized optimal solutions provided by a critic or the dynamic relative performance within a group, and establishes a more general advantage baseline.

% From the perspective of human learning, we view rollout as repeated practice, and this grouping policy is similar to Trial-and-Error, in which humans often compare the results with a fixed \textbf{Reference Anchor} after repeated practice in order to improve the efficiency of Trial-and-Error learning. This anchor serves as an objective frame of reference, different from both idealized optimal solutions provided by Critic and dynamically changing relative performance within a group, and it provides a more universal advantage baselines.

%our tech Pre-Estimated Value-Based % Precomputed Value 
In this paper, we introduce Pre-estimated Value-based Policy Optimization (PVPO), a generalized RL method based on Proximal Policy Optimization (PPO) \citep{schulman2017proximalpolicyoptimizationalgorithms}. PVPO adopts a critic-free architecture, is compatible with mainstream group policy RL methods, and maintains low computational cost for grouping, thus effectively combining the strengths of both approaches.
Specifically, we use a Reference Model (Ref) to run grouping reasoning and calculate a task-based reward score as an anchor. This anchor serves as the $V$ estimate during RL training, helping to correct the cumulative bias in relative advantage calculations typically observed in large language models (LLMs).
In essence, our method decouples $Q$ and $V$ in the grouping policy advantage calculation. The reference anchor is computed in an unsupervised manner and acts as both a supplement and an enhancement to the training dataset, without incurring additional time or memory overhead.
In summary, our core contributions are as follows.

% In this study, we introduce Pre-estimated Value-based Policy Optimization (PVPO), a generalized RL method that is a variant of Proximal Policy Optimization (PPO) \citep{schulman2017proximalpolicyoptimizationalgorithms} that follows the Critic-Free architecture while being compatible with mainstream group policy RL methods, and is able to use minimal grouping computational cost, fully combining the benefits of both RL architectures.
% Specifically, we perform a grouping reasoning using the Reference Model (Ref), and calculate the reward score based on task as an anchor, which participates in advantage computation as a V estimate during the RL training stage, thus correcting the cumulative bias of large language model (LLM) for calculating the relative advantage within group. 
% In essence, we decouple the Q and V for advantage computation in the grouping policy, 
% Note that the \textbf{Reference Anchor} computation is unsupervised and serves as a complement and enhancement to the training dataset without increasing the time and space consumption.

\begin{itemize}
    \item 
    We propose PVPO, an efficient and generalizable approach to critic-free reinforcement learning.
    PVPO provides a stable, low-variance, and globally consistent advantage function, effectively mitigating concerns of error accumulation and policy drift during training. As a result, PVPO enables more efficient and robust policy optimization while significantly reducing spatio-temporal overhead.
    \item 
    We introduce a group sampling strategy that offline filters data with unstable accuracy rates to construct high-quality batches, thereby enhancing convergence and learning efficiency. Furthermore, for samples with zero accuracy (i.e., zero reward), we leverage a large-scale LLM to generate ground-truth trajectories, facilitating more effective learning from sparse reward signals. %, where positive feedback only occurs occasionally, such as when the agent finds the correct answer, and most rollouts receive no reward. %Furthermore, for samples with zero accuracy, we leverage a large-scale LLM to generate ground-truth trajectories, facilitating more effective learning from sparse reward signals.
    \item 
    PVPO achieves state-of-the-art performance on multi-step retrieval datasets and demonstrates strong generalization on mathematical reasoning benchmarks. Experimental results indicate that PVPO not only enhances multi-hop question answering (QA) and tool-use capabilities, but also improves the overall reasoning ability of LLMs.
\end{itemize}

\section{Related Work}
% TODO: Put Related work before or after Conclusion.
% \subsection{Agentic Search}
\subsection{Agentic Reasoning}
Leveraging reinforcement learning to drive search represents an important direction in agentic reasoning~\citep{jin2025searchr1trainingllmsreason, jiang2025deepretrievalhackingrealsearch}.
Search-o1~\citep{li2025searcho1agenticsearchenhancedlarge} integrates an agentic search workflow into the reasoning trajectory. This achieves an elegant integration of search and reasoning, sparking a wave of subsequent optimizations \citep{qian2025toolrlrewardtoollearning, wang2025stepsearchignitingllmssearch,feng2025retoolreinforcementlearningstrategic}.
Moreover, numerous studies on Retrieval-Augmented Generation (RAG) \citep{li2025searcho1agenticsearchenhancedlarge, feng2025airrag, hao2025dynasearcherdynamicknowledgegraph} have advanced the capabilities of LLM in tool use and information retrieval.

However, existing studies often directly apply algorithms such as GRPO, which are intrinsically ill-suited to the sparse-reward setting of agentic search. These methods depend on dense token-level rewards, necessitating extensive rollouts to achieve stable advantage estimation. Consequently, the quality of the learning signal becomes tightly coupled with the sample size.

Our PVPO framework is tailored for agentic search by decoupling the advantage function ($A$=$Q$-$V$), thereby mitigating sample size dependency. While the actual return ($Q$) leverages the sample size, the advantage baseline ($V$) remains independent of both the current and previous policies. This design ensures a stable learning signal even under severe reward sparsity (e.g., $Q$=0), obviating the need for extensive rollouts.

% Our PVPO is designed for agentic search by decoupling the advantage function (A = Q - V) to break the dependency on sample size.
% The actual return (Q) benefits from the sample size and the advantage baseline (V) is independent of the current or old policy.
% This design provides a stable learning signal even under extreme reward sparsity (Q=0), eliminating the need for extensive rollouts.

% \subsection{Mathematical Reasoning}
% TODO: This section discusses related work on the integration of mathematical reasoning and reinforcement learning.

% \subsection{Reward Design and Advantage Computation}
\subsection{RL for LLMs}
Recently, reward and advantage computation has been redefined through dynamic generation and iterative optimization, substantially enhancing the performance of critic-free RL methods. Some methods construct denser feedback signals by increasing the frequency of reward generation~\citep{bensal2025reflectretryrewardselfimproving, chen2024selfplayfinetuningconvertsweak}, while others improve reward adherence by incorporating additional training phases into the learning process~\citep{dong2025toolstarempoweringllmbrainedmultitool}. These approaches often overlook the compounding hallucinations arising from repeated sampling and error accumulation from iterative policy updates. Each incremental policy change alters the rollout distribution, resulting in advantage estimates targeting a continually shifting objective and potentially steering the policy toward suboptimal local minima. Moreover, these methods depend heavily on costly online sampling procedures.

Another line of research seeks to recover endogenous rewards from the actor model via reverse engineering, a process that has been mathematically substantiated~\citep{li2025generalistrewardmodelsinside, zhao2025learningreasonexternalrewards}. This approach eliminates the need for additional training and enables adaptation to diverse evaluation preferences through prompt adjustment. However, the quality of the recovered reward is inherently limited by the base model’s capabilities, and consistently guiding reward signals through prompting remains a significant challenge~\citep{zhao2021calibrateuseimprovingfewshot, lu2022fantasticallyorderedpromptsthem, liu2021pretrainpromptpredictsystematic}.

% Another research idea is to recover the endogenous reward from the actor model through reverse engineering, which has already provided a rigorous mathematical proof \citep{li2025generalistrewardmodelsinside, zhao2025learningreasonexternalrewards}. 
% This reward requires no additional training and can be adapted to different evaluation preferences by adjusting the prompts.
% But the reward quality ceiling is bounded by the capabilities of the base model. It is extremely challenging to guide reward signals consistently through prompting alone \citep{zhao2021calibrateuseimprovingfewshot, lu2022fantasticallyorderedpromptsthem, liu2021pretrainpromptpredictsystematic}.

To address these challenges, the research community has investigated various static approaches. The most prominent is offline reinforcement learning, which optimizes policies using fixed datasets \citep{kumar2020conservativeqlearningofflinereinforcement, kostrikov2021offlinereinforcementlearningimplicit}. Another notable class comprises Direct Preference Optimization (DPO) \citep{rafailov2024directpreferenceoptimizationlanguage} and its variants \citep{ethayarajh2024ktomodelalignmentprospect}, which reformulate the objective as a direct fit to fixed preference pairs, reducing the reliance on online sampling but constraining generalization. Simpler static methods, such as weighted behavioral cloning \citep{xu2022policyguided, xu2022discriminatorweightedofflineimitationlearning}, offer limited expressive power and theoretical guarantees due to their parsimonious advantage estimation.

% To alleviate these issues, the research community has explored a number of static approaches.
% The most dominant of these is offline RL, which learns policy entirely on a fixed dataset \citep{kumar2020conservativeqlearningofflinereinforcement, kostrikov2021offlinereinforcementlearningimplicit}.
% Another class of static methods is Direct Preference Optimization (DPO) \citep{rafailov2024directpreferenceoptimizationlanguage} and its variants \citep{ethayarajh2024ktomodelalignmentprospect}. This class of methods cleverly transforms the optimization objective into a direct fit to fixed preference pairs, which avoids a lot of online sampling but limits the generalization.
% Simpler static methods, such as weighted behavioral cloning \citep{xu2022policyguided, xu2022discriminatorweightedofflineimitationlearning}, are too parsimonious in advantage estimation to provide strong expressive power and theoretical support.

To balance efficiency and adaptability in policy optimization, our approach integrates a static $V$ with a dynamic $Q$, ensuring stable advantage estimation and low computational overhead while maintaining responsive adaptation to policy updates.

\section{Preliminary}
In this section, we review the fundamental concepts of policy optimization in RL, with a particular focus on the role of the advantage function and its various estimation methods.

\subsection{Proximal Policy Optimization}
Actor-critic methods, such as PPO, train a critic network $V_\phi(s)$  to provide a low-variance estimate of the state-value function $V^\pi(s)$ of state $s$. 
The state-value function is used to compute the advantage at each time step $t$, typically via Generalized Advantage Estimation (GAE) \citep{schulman2018GAE}:
\begin{equation}
\begin{split}
\hat{A}^{\text{GAE}}_t  = \sum_{l=0}^{\infty} (\gamma\lambda)^l \delta_{t+l} , \quad \delta_t = r_t + \gamma V_\phi(s_{t+1}) - V_\phi(s_t) \text{,}
\end{split}
\end{equation}
%where $\delta_t = r_t + \gamma V_\phi(s_{t+1}) - V_\phi(s_t)$.
where $\lambda$ is a hyper-parameter, $\delta_t$ is the temporal difference error at time step $t$, $r_t$ is the immediate reward received at time step $t$, $\gamma$ is the discount factor.
PPO then optimizes a clipped surrogate objective to update the actor network in a stable manner:
\begin{equation}
\begin{split}
\mathcal{J}^{\text{PPO}}(\theta) ={}& \mathbb{E}_{q\sim P(D), o\sim \pi_{\theta_{\text{old}}}(O|q)} \left[ \min \left( r_t(\theta) \hat{A}^{\text{GAE}}_t, \text{clip}(r_t(\theta), 1 - \epsilon, 1 + \epsilon) \hat{A}^{\text{GAE}}_t \right) \right] \text{,}
\end{split}
\end{equation}
%where $r_t(\theta)$ is the probability ratio $\frac{\pi_\theta(a_t|s_t)}{\pi_{\theta_{\text{old}}}(a_t|s_t)}$. 
where $q$ are questions sampled from the dataset $D$, $o$ are outputs sampled from the old policy $\pi_{\text{old}}$, importance sampling ratio $r_{t}(\theta)=\frac{\pi_\theta(o_{t}|q, o_{<t})}{\pi_{\theta_{\text{old}}}(o_{t}|q, o_{<t})}$, $\epsilon$ is the clipping range of $r_{t}(\theta)$.

\subsection{Group Relative Policy Optimization}
Since the critic network is typically as large as the actor network, it adds substantial memory and computational burden. 
Critic-free methods, such as GRPO, eliminate this costly component by estimating the advantage directly from rewards.

For each question, GRPO generates a group of outputs $\{o_i\}$ from the old policy $\pi_{\theta_{\text{old}}}$. The advantage for each output $o_i$ is then calculated based on normalized reward $\mathbf{r}$ relative to the group:
\begin{equation}
\hat{A}_{i,t} = \frac{r_i - \text{mean}(\mathbf{r})}{\text{std}(\mathbf{r})} \text{.}
\end{equation}
This critic-free advantage estimate is then used to optimize a PPO-like objective function:
%\begin{equation}
%J^{\text{GRPO}}(\theta) = \mathbb{E}_{q, \{o_i\}} \left[ \frac{1}{G} \sum_{i=1}^{G} \frac{1}{|o_i|} \sum_{t=1}^{|o_i|} \left\{ \min \left( \frac{\pi_\theta(o_{i,t}|q, o_{i,<t})}{\pi_{\theta_{\text{old}}}(o_{i,t}|q, o_{i,<t})} \hat{A}_{i,t} , \text{clip} \left( \frac{\pi_\theta(o_{i,t}|q, o_{i,<t})}{\pi_{\theta_{\text{old}}}(o_{i,t}|q, o_{i,<t})}, 1-\epsilon, 1+\epsilon \right) \hat{A}_{i,t} \right) - \beta D_{KL}[\pi_\theta || \pi_{\text{ref}}] \right\} \right]
%\end{equation}
\begin{equation}
\begin{split}
& \mathcal{J}^{\text{GRPO}}(\theta)  = \mathbb{E}_{q\sim P(D), \{o_i\}\sim \pi_{\theta_{\text{old}}}(O|q)}  \\
&\quad \Biggl[ \frac{1}{G} \sum_{i=1}^{G} \frac{1}{|o_i|} \sum_{t=1}^{|o_i|} \biggl\{ \min \left( r_{i,t}(\theta) \hat{A}_{i,t} , \text{clip} \left( r_{i,t}(\theta), 1-\epsilon, 1+\epsilon \right) \hat{A}_{i,t} \right) - \beta D_{KL}[\pi_\theta || \pi_{\text{ref}}] \biggr\} \Biggr] \text{,}
\end{split}
\end{equation}
where $r_{i,t}(\theta)=\frac{\pi_\theta(o_{i,t}|q, o_{i,<t})}{\pi_{\theta_{\text{old}}}(o_{i,t}|q, o_{i,<t})}$, $D_{KL}$ is the KL divergence between the trained policy $\pi_\theta$ and the reference policy $\pi_{\text{ref}}$, $\beta$ is a hyper-parameter.

\begin{figure*}[t]
    \centering 
    \includegraphics[width=\textwidth]{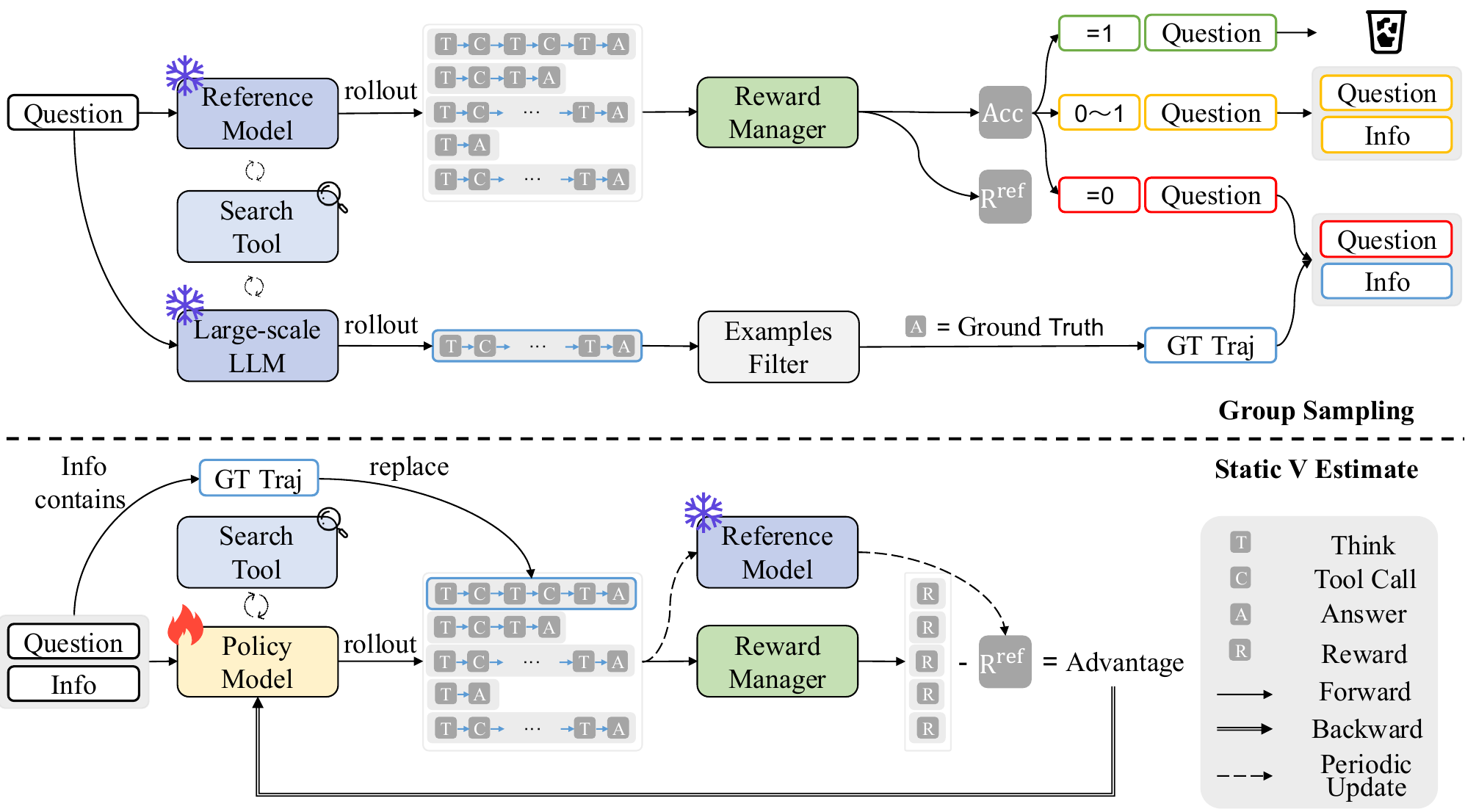} 
    \caption{The architecture of PVPO. Reference model updates $R^{\text{ref}}$ at fixed steps, maintaining value stability and improving the performance lower bound. Reward manager do not restrict the generation of reward.}
    \label{fig:PVPO}
\end{figure*}

\section{Methodology}
% Pre-Estimated Value-Based % Precomputed Value 
In this section, we will introduce our efficient and effective RL algorithm PVPO. 
The architecture is illustrated in Figure \ref{fig:PVPO}.
PVPO optimizes the policy via the following objective:

\begin{equation}
\begin{split}
& \mathcal{J}^{\text{PVPO}}(\theta)  = \mathbb{E}_{q\sim P(D), \{o_i\}\sim \pi_{\theta_{\text{old}}}(O|q)} \\
&\quad \Biggl[ \frac{1}{G} \sum_{i=1}^{G} \frac{1}{|o_i|} \sum_{t=1}^{|o_i|}  \biggl\{ \min \left( r_{i,t}(\theta) \hat{A}^{\text{PVPO}}_{i,t} , \text{clip} \left( r_{i,t}(\theta), 1-\epsilon, 1+\epsilon \right) \hat{A}^{\text{PVPO}}_{i,t} \right) - \beta D_{KL}[\pi_\theta || \pi_{\text{ref}}] \biggr\} \Biggr] \text{.}
\end{split}
\end{equation}
where 
\begin{equation}
r_{i,t}(\theta) =
\begin{cases}
    \frac{\pi_\theta(o_{i,t}|q, o_{i,<t})}{\pi_{\theta_{\text{old}}}(o_{i,t}|q, o_{i,<t})}, & \text{if } o_i \notin \text{GT Traj} .\\
    \frac{\pi_\theta(o_{i,t}|q, o_{i,<t})}{\pi_{\theta_{\text{gt}}}(o_{i,t}|q, o_{i,<t})}, & \text{if } o_i \in \text{GT Traj} .
\end{cases}
\end{equation}

\subsection{Static V Estimate}
In actual policy optimization, the current method is to operate at the group level rather than through single sampling. 
For problem $q$, we use the current policy $\pi_\theta$ to generate $N$ independent trajectories  $\mathcal{T}=\{\tau_1, \tau_2, ..., \tau_N\}$ and obtain the corresponding rewards $\mathbf{r} =\{R(\tau_1) ,R(\tau_2) ,...,R(\tau_N) \} =\{r_1, r_2, ..., r_N\}$.
For any step $(s_{i,t}, a_{i,t})$ in a specific trajectory $\tau_i$, the unbiased Monte Carlo estimate of the action value $Q^\pi(s_{i,t}, a_{i,t})$ is the final reward $r_i$ observed in that trajectory. 
We refer to this as the \textbf{Dynamic Q Estimate} because it directly reflects the result of a single rollout of the current policy:
\begin{equation}
\label{dynQ}
\hat{Q}_{\text{dyn}}(\tau_i) = \mathbb{E}_{\tau \sim \pi_{\theta}} [R(\tau_i)] = r_i \text{.}
%, \text{for all } (s_{i,t}, a_{i,t}) \in \tau_i
\end{equation}
Considering that reward $r_i$ is given after the generation of trajectory $\tau_i$,
the trajectory generation process is regarded as atomic actions $a_i = \tau_i$ executed from $s_{i,0}$.
This atomicity makes the reward distribution of the intermediate state $s_{i,t}$ only depend on initial state $s_{i,0}$ ($s_0$) and $\pi_i$.
Consequently, the expected return of the policy is equal to the state value of the initial state $V^\pi(s_0)$.
A natural estimation method is to approximate this expectation using the empirical mean of all rewards in the current group. 
This is the approach adopted by on-policy methods such as GRPO, which we refer to as \textbf{Dynamic V Estimate}:
\begin{equation}
\label{dynV}
\hat{V}_{\text{dyn}}(s_0)=\hat{V}_{\text{dyn}}(\mathcal{T}) = \frac{1}{N} \sum_{j=1}^{N} r_j = \text{mean}(\mathbf{r}) \text{.}
\end{equation}
So we obtain the sparse advantage estimate for trajectory $\tau_i$ in the on-policy method:
\begin{equation}
\begin{split}
% \hat{A}_{\text{dyn}}(\tau_i,s_0) & = \hat{Q}_{\text{dyn}}(\tau_i) - \hat{V}_{\text{dyn}}(s_0) \\
% & = r_i - \text{mean}(\mathbf{r}) \text{.}
\hat{A}_{\text{dyn}}(\tau_i,s_0) & = \hat{Q}_{\text{dyn}}(\tau_i) - \hat{V}_{\text{dyn}}(s_0) = r_i - \text{mean}(\mathbf{r}) \text{.}
\end{split}
\end{equation}
This formula clearly shows that the advantage is calculated as the difference between the immediate reward and the average performance of the current policy $\pi_\theta$ within the group. 
However, $\hat{V}_{\text{dyn}}$ fluctuates wildly with each sampling of the group and is directly affected by $\pi_\theta$, introducing significant instability, especially when the group size is not large enough. To more effectively mitigate the instability associated with dynamic $V$ estimation, we propose substituting it with a more robust fixed $V$ estimate.

The ideal baseline should represent a \textbf{Reference Anchor} that does not change with current policy iterations.
Therefore, we use the expected return of a fixed reference policy $\pi_{ref}$ (e.g., the initial policy model) as our \textbf{Static V Estimate} $\hat{V}_{\text{sta}}$. The baseline can be accurately estimated in advance by sampling the reference policy $\pi_{ref}$ M times, and update at fixed steps during training process:
\begin{equation}
%\mathbb{E}_{\tau \sim \pi_{ref}} [R(\tau) | s_0]  \approx 
\hat{V}_{\text{sta}}(s_0) =\frac{1}{M} \sum_{j=1}^{M} r^{\text{ref}}_j =  \text{mean}(\mathbf{r^{\text{ref}}})\text{.}
\end{equation}
This stable static baseline replaces the unstable dynamic baseline in formula \ref{dynV}. We finally obtain the advantage function of PVPO, which is well-suited for RL tasks with sparse rewards. %which is applicable to all sparse reward fields:
\begin{equation}
\begin{split}
% \hat{A}^{\text{PVPO}}(\tau_i, s_0) & = \hat{Q}_{\text{dyn}}(\tau_i) - \hat{V}_{\text{sta}}(s_0) \\
% & = r_i - \text{mean}(\mathbf{r^{\text{ref}}}) \text{.}
\hat{A}^{\text{PVPO}}(\tau_i, s_0) & = \hat{Q}_{\text{dyn}}(\tau_i) - \hat{V}_{\text{sta}}(s_0) = r_i - \text{mean}(\mathbf{r^{\text{ref}}}) \text{.}
\end{split}
\end{equation}

% For any step $(s_{i,t}, a_{i,t})$ of any trajectory $\tau_i$, the advantage can also be written as follows:
% \begin{align}
%     \hat{A}^{\text{PVPO}}_{i,t} &= r_i - \text{mean}(\mathbf{r}^{\text{ref}})\text{,}  \\ % 
%     r_{i} &=
%     \begin{cases}
%         \max(0.1,\, r_{\text{acc}}), & \text{if format is correct.} \\
%         0, & \text{if format is incorrect.}
%     \end{cases}  \\ %  \nonumber 
%     r_{\text{acc}} &=
%     \begin{cases}
%         \text{F}_{1}(a_{\text{pred}},\, a_{\text{gt}}), & \text{if } L_{\text{pred}} \geq n \cdot L_{\text{gt}}. \\
%         \text{CEM}(a_{\text{pred}},\, a_{\text{gt}}), & \text{if } L_{\text{pred}} < n \cdot L_{\text{gt}}.
%     \end{cases}  % 
% \end{align}
% where $a_{*}$ denotes the answer, $L_{*}$ denotes the length of the answer, $*_{pred}$ denotes the prediction, and $*_{gt}$ denotes the ground truth. $r^{ref}_i$ is similar.
% F$_1$ denotes the standard word-level F$_1$ score and CEM denotes Cover Exact Match. 
% $n$ represents the multiple of the text length, which we set to $3$ by default. 
% This reward is used for agentic search, following \cite{hao2025dynasearcherdynamicknowledgegraph}.
% This reward is set for the field of agentic search. 
% This type of reward is a common setting in the field of agentic reasoning.

In summary, $\hat{Q}_{\text{dyn}}(\tau_i)$ is obtained from the immediate reward of on-policy $\pi_\theta$ rollout. It reflects the current performance of the policy and is highly adaptive. 
The \textbf{Static V Estimate} $\hat{V}_{\text{sta}}(s_0)$ is obtained from the average reward of the reference policy $\pi_{\text{ref}}$ pre-rollout. 
It provides a stable and low-variance performance baseline.

\subsection{Group Sampling}
% Inspired by DAPO's dynamic sampling method, we also calculate the accuracy of the sample rollout, but we still use the reference model for offline rollouts. 
% For each sample, we select the mean accuracy of the rollouts as the filtering standard. 

% Specifically, we divide the samples into three groups.
% Samples with a mean accuracy of 1 are directly removed from the training set, as such samples are too simple to improve the model's learning ability.
% Samples with a mean accuracy between 0 and 1 are retained, because the resulting advantage for this group is not zero.

% For samples with a mean accuracy of 0, we first perform an additional rollout using a big-scale LLM.
% The big-scale LLM can correctly answer some samples, and we cache these Grund Truth Trajectories (GT Traj) and the corresponding probability distributions.
% A GT Traj is injected by replacing one of generated rollouts for these specific samples during policy training.
% The underlying principle is to address the sparse reward problem inherent in complex samples. Without guidance, the LLM may fail to discover any positive feedback through unguided exploration. A reference trajectory acts as a demonstration, bootstrapping the learning process by providing a clear and structured signal of a successful reasoning path.

% Inspired by DAPO's dynamic sampling method, we also calculate the accuracy of sample rollouts, but we still use the reference model for offline rollouts. For each sample, we use the mean accuracy of the rollouts as the filtering standard.

Inspired by DAPO's dynamic sampling strategy, we also assess the accuracy of sample rollouts while continuing to utilize the reference model for offline rollouts. For each sample, the mean accuracy of the rollouts serves as the filtering criterion.

Specifically, samples are categorized into three groups:
% \begin{itemize}
% \item Samples with a mean accuracy of 1 are excluded from the training set, as they are considered too trivial to facilitate effective learning.
% \item Samples with a mean accuracy strictly between 0 and 1 are retained, given their nonzero advantage.
% \item For samples exhibiting a mean accuracy of 0, an additional rollout is conducted using a larger LLM for further evaluation.
% \end{itemize}
% \begin{itemize}[noitemsep,topsep=0pt,parsep=0pt,partopsep=0pt]
\begin{itemize}[noitemsep,topsep=0pt,parsep=0pt,partopsep=0pt]
    \item Samples with a mean accuracy of 1 are excluded from the training set, as they are considered too trivial to facilitate effective learning.
    \item Samples with a mean accuracy strictly between 0 and 1 are retained, given their nonzero advantage.
    \item For samples exhibiting a mean accuracy of 0, an additional rollout is conducted using a larger LLM for further evaluation.
\end{itemize}

The larger LLM can correctly answer some of these samples. We cache these Ground Truth Trajectories (GT Traj) and their probability distributions. During policy training, a GT Traj is injected by replacing one of the generated rollouts for these specific samples. 
% This method helps address the sparse reward problem in complex samples. Without guidance, the LLM might not find any positive feedback through unguided exploration. A reference trajectory provides a demonstration, jumpstarting learning by showing a clear example of a successful reasoning process.
This method mitigates the sparse reward issue commonly encountered with complex samples. In the absence of guidance, the LLM may fail to obtain any positive feedback through unguided exploration. By providing a reference trajectory, the model receives an explicit demonstration, which jumpstarts learning by offering a clear example of a successful reasoning process.

\section{Expriments Setting}
\textbf{Metrics.} 
For multi-hop QA tasks, we employ answer accuracy (Acc, \%) and LLM-as-a-Judge (LasJ, \%)~\citep{song2025r1searcherincentivizingdynamicknowledge} as evaluation metrics. 
For mathematical reasoning tasks, we measure answer accuracy (Acc, \%), reporting the mean accuracy across 32 independent rollouts for each sample (i.e., acc@32).

\textbf{Datasets.}
For multi-hop QA tasks, we conduct experiments on four multi-step retrieval datasets: Musique~\citep{trivedi2022musiquemultihopquestionssinglehop}, 2WikiMultiHopQA (2Wiki)~\citep{ho2020constructingmultihopqadataset}, HotpotQA~\citep{yang2018hotpotqadatasetdiverseexplainable}, and Bamboogle (Bam)~\citep{press2023measuringnarrowingcompositionalitygap}.
Model training is performed on the Musique training split, which consists of 20k examples, and evaluations are carried out on the full development and test sets. 
For mathematical reasoning tasks, we train models on DAPO-Math-17k-Processed~\citep{yu2025dapoopensourcellmreinforcement}, comprising 17k examples, and conduct evaluation on five test sets: DAPO-AIME-2024~\citep{AIMO2024,BytedTsinghua-SIA_AIME_2024}, AIME-2025~\citep{lin2024aime2025}, MATH500~\citep{lightman2023lets,huggingface_math500}, AMC23~\citep{aimo2024validation_amc}, and Olympiad~\citep{he2024olympiadbench}.

\textbf{Baselines and Training Details.}
We use \textit{Qwen2.5-7B-Instruct} and \textit{Qwen2.5-14B-Instruct} as base models and \textit{Qwen2.5-72B-Instruct} as the large LLM to generate GT Traj. The reference reward $R^{\text{ref}}$ is updated every 500 steps. 
For training, we set the learning rate to 1e-6, maximum response length to 8192, sampling temperature to 1.0 and top-p to 1.0. For inference, we set the sampling temperature to 0.6 and top-p to 0.95.
For the multi-hop QA tasks, we benchmark our method against not only state-of-the-art LLMs such as \textit{DeepSeek-R1-0528}, \textit{GPT-4.1-0414}, \textit{O4-mini-0416}, and \textit{Gemini-2.5-pro-0325}, but also prominent RL-based agentic search models~\citep{jin2025searchr1trainingllmsreason,song2025r1searcherincentivizingdynamicknowledge}.
We adopt the ReSearch~\citep{chen2025research} framework, with pre-samples $M=5$, rollout $N=5$, train batch size of 8, and 1,000 training steps. For DynaSearcher\citep{hao2025dynasearcherdynamicknowledgegraph}, we remove the ``kg\_filter" during inference. %More details can be found in Appendix~\ref{impl_detail}.
For mathematical reasoning tasks, we primarily adopt GRPO~\citep{shao2024deepseekmathpushinglimitsmathematical}, DAPO~\citep{yu2025dapoopensourcellmreinforcement}, and GSPO~\citep{zheng2025groupsequencepolicyoptimization} as baselines. 
We use the verl~\citep{sheng2024hybridflow} framework with pre-samples $M=16$, rollout $N=16$, train batch size of 32, and 1,000 training steps.
For DAPO, we set the clipping parameter $\epsilon_{\text{low}}=0.2$ and $\epsilon_{\text{high}}=0.28$. 
% For GRPO, we configured the ``loss\_agg\_mode" to ``token-mean" instead of ``seq-mean-token-mean". 
For GRPO, we set the ``loss\_agg\_mode" to ``seq-mean-token-mean", which is aligned with the original paper.
For GSPO, the clipping parameter $\epsilon$ is set to 0.0003. 
All experiments are conducted on a server equipped with an Intel(R) Xeon(R) Platinum 8369B CPU and $8\times$NVIDIA A100-SXM4-80GB GPUs. More details can be found in Appendix~\ref{impl_detail}.

\section{Experiments}
In this section, we conduct a series of experiments to comprehensively evaluate PVPO. First, we test our method on multi-hop QA to validate its effectiveness in the agent domain. Next, we perform ablation studies to examine the contributions of the core modules of PVPO. We further apply PVPO to mathematical reasoning tasks to verify its generalizability and also evaluate its compatibility with other advanced RL algorithms. In addition, we analyze the training efficiency and convergence properties of PVPO. Finally, we present a case study to investigate the efficiency and robustness of PVPO under low sampling budget.

\begin{table*}[tbp] % [htbp] 
  \centering 
  \setlength{\tabcolsep}{3pt}
  \caption{Performance comparisons between PVPO and the baselines on multi-step retrieval datasets. The best and second best results are \textbf{bold} and \underline{underlined}, respectively. %$\dag/\ddag$ represents in-domain/out-of-domain datasets.
  }
  \label{tab:QA_performance} 
  \begin{tabular}{l cccccccccc} 
    \toprule 
   
    % \multirow{2}{*}{Method} & \multicolumn{2}{c}{\textbf{Musique}\dag} & \multicolumn{2}{c}{\textbf{2Wiki}\dag} & \multicolumn{2}{c}{\textbf{HotpotQA}\dag} & \multicolumn{2}{c}{\textbf{Bamboogle}\ddag} & \multicolumn{2}{c}{\textbf{Average}} \\
    \multirow{2}{*}{Method} & \multicolumn{2}{c}{\textbf{Musique}} & \multicolumn{2}{c}{\textbf{2Wiki}} & \multicolumn{2}{c}{\textbf{HotpotQA}} & \multicolumn{2}{c}{\textbf{Bamboogle}} & \multicolumn{2}{c}{\textbf{Average}} \\
    
    \cmidrule(r){2-3} \cmidrule(r){4-5} \cmidrule(r){6-7} \cmidrule(r){8-9} \cmidrule(r){10-11} & \textbf{Acc} & \textbf{LasJ} & \textbf{Acc} & \textbf{LasJ} & \textbf{Acc} & \textbf{LasJ} & \textbf{Acc} & \textbf{LasJ} & \textbf{Acc} & \textbf{LasJ} \\
    \hline %\midrule %
    \rowcolor{tablegray}
    \multicolumn{11}{c}{\textit{\textbf{Prompt Based}}} \\
    \hline %\midrule %
    Qwen2.5-7B-Instruct      & 5.1  &13.5 & 27.9 &29.3 & 22.4 &31.0 & 12.8 &17.1 & 17.1 &22.7  \\
    DeepSeek-R1              & 32.0 &40.7 & 57.5 &59.4 & 43.0 &58.3 & 66.4 &76.6 & 49.7 &58.8 \\
    O4-mini                  & 38.0 &44.1 & 61.5 &67.4 & 49.5 &67.4 & \underline{74.4} &\underline{84.2} & 55.9 &65.8 \\
    GPT-4.1-global           & 31.0 &40.9 & 58.0 &58.5 & 44.5 &57.7 & 51.2 &61.6 & 46.2 &54.7 \\
    Gemini-2.5-pro           & \underline{42.5} &50.8 & 70.0 &71.2 & 53.0 &71.1 & \textbf{75.2} &\textbf{84.5} & \underline{60.2} &\underline{69.4} \\
    
    % Qwen2.5-7B-Instruct      & 5.10  &13.48 & 27.90 &29.30 & 22.40 &31.04 & 12.80 &17.12 & 17.05 &22.74  \\
    % DeepSeek-R1              & 32.00 &40.70 & 57.50 &59.40 & 43.00 &58.30 & 66.40 &76.64 & 49.73 &58.76 \\
    % O4-mini                  & 38.00 &44.10 & 61.50 &67.40 & 49.50 &67.40 & 74.40 &84.16 & 55.85 &65.77 \\
    % GPT-4.1-global           & 31.00 &40.90 & 58.00 &58.50 & 44.50 &57.70 & 51.20 &61.60 & 46.18 &54.68 \\
    % Gemini-2.5-pro           & 42.50 &50.80 & 70.00 &71.20 & 53.00 &71.10 & 75.20 &84.48 & 60.18 &69.40 \\
    %PVPO-Qwen2.5-7B-Instruct  & Post-Train              & 45.30  & 78.80  & 68.20  & 48.00  & 60.08  \\
    \hline %\midrule %
    \rowcolor{tablegray}
    \multicolumn{11}{c}{\textit{\textbf{Train Based}}} \\
    \hline %\midrule %
    \multicolumn{7}{l}{\textit{\textbf{Qwen2.5-7B-Instruct}}} \\
    % ReSearch              & 13.40 &24.90 & 41.70 &43.26 & 33.40 &47.26 & 36.00 &43.20 & 31.13 &39.66 \\
    % GRPO on ReSearch      & 33.40 &46.72 & 60.80 &67.02 & 54.50 &63.68 & 45.60 &54.40 & 48.58 &57.96 \\
    % GRPO on DynaSearcher  & \underline{38.90} &\underline{52.04} & \underline{74.30} &\underline{76.77} & 62.70 &68.32 & \textbf{51.20} &\underline{58.72} & \underline{56.78} &\underline{63.96} \\
    % PVPO on ReSearch      & 36.50 &51.44 & 70.10 &72.36 & \underline{65.50} &\underline{72.34} & 45.60 &54.32 & 54.43 &62.62 \\
    % PVPO on DynaSearcher  & \textbf{46.90} &\textbf{59.44} & \textbf{77.70 } &\textbf{80.62} & \textbf{69.00}  &\textbf{78.44} & \underline{50.40} &\textbf{59.68} & \textbf{61.00} &\textbf{69.55} \\

    Search-R1-v0.3  & 24.7 &34.6 & 58.7 &61.1 & 53.6 &66.9 & 48.0 &54.5 & 46.3 &54.4 \\
    R1-Searcher     & 24.7 &34.2 &67.8 & 68.2 &59.7 & 71.5 &46.4 & 52.0 &50.5 & 56.5 \\
    GRPO-ReSearch        & 33.4 &46.7 & 60.8 &67.0 & 54.5 &63.7 & 45.6 &54.4 & 48.6 &58.0 \\
    % GRPO-DynaSearcher    & \underline{38.9} &\underline{52.0} & \underline{74.3} &\underline{76.8} & 62.7 &68.3 & \textbf{51.2} &\underline{58.7} & \underline{56.8} &\underline{64.0} \\
    % PVPO-ReSearch   & 36.5 &51.4 & 70.1 &72.4 & \underline{65.5} &\underline{72.3} & 45.6 &54.3 & 54.4 &62.6 \\
    % PVPO-DynaSearcher  & \textbf{46.9} &\textbf{59.4} & \textbf{77.7 } &\textbf{80.6} & \textbf{69.0}  &\textbf{78.4} & \underline{50.4} &\textbf{59.7} & \textbf{61.0} &\textbf{69.6} \\
    GRPO-DynaSearcher    & 38.9 & \underline{52.0} & \underline{74.3} & \underline{76.8} & 62.7 &68.3 & 51.2 & 58.7 & 56.8 & 64.0 \\
    PVPO-ReSearch   & 36.5 &51.4 & 70.1 &72.4 & \underline{65.5} & \underline{72.3} & 45.6 &54.3 & 54.4 &62.6 \\
    PVPO-DynaSearcher  & \textbf{46.9} &\textbf{59.4} & \textbf{77.7 } &\textbf{80.6} & \textbf{69.0}  &\textbf{78.4} & 50.4 & 59.7 & \textbf{61.0} &\textbf{69.6} \\
    \bottomrule 
  \end{tabular}
\end{table*}

\subsection{Main Results}
% To answer \textbf{Q1}, we evaluated PVPO on multi-hop QA datasets and compared the results with two types of benchmarks. The results are shown in Table~\ref{tab:QA_performance}.
% We evaluated PVPO against both the zero-shot inference performance of leading LLMs and various trained RL-based search methods. The results, presented in Table~\ref{tab:QA_performance}, underscore the effectiveness of our approach.
% Specifically, applying PVPO elevates the performance of the base frameworks. For ReSearch, it achieves improvements of 5.8 points in Acc and 4.6 points in LasJ. For DynaSearcher, the gains are 4.2 points in Acc and 5.6 points in LasJ.
% Notably, our PVPO-DynaSearcher significantly outperforms all RL-trained baselines, marginally surpasses the strongest proprietary LLM, \textit{Gemini-2.5-Pro} and establishes a considerable lead over other state-of-the-art models, including \textit{O4-mini}, \textit{GPT-4.1}, and \textit{DeepSeek-R1}.
% We observed that the performance of PVPO improved by more than 5 points on average compared to GRPO.
% The result demonstrates the consistent superiority of PVPO across agentic retrieval methods.
% Overall, this answers \textbf{Q1}: Compared to existing reinforcement learning methods, PVPO achieves state-of-the-art performance.
We evaluate PVPO against both zero-shot leading LLMs and trained RL-based search methods, with results in Table~\ref{tab:QA_performance} underscoring its effectiveness. Specifically, applying PVPO substantially improves the base frameworks, boosting ReSearch's Avg Acc/LasJ scores by 5.8/4.6 points and DynaSearcher's by 4.2/5.6 points.
Notably, our PVPO-DynaSearcher model significantly outperforms all RL-trained baselines (e.g., surpassing GRPO by over 5 points on average). It also marginally exceeding the strongest proprietary LLM, \textit{Gemini-2.5-Pro}, while establishing a considerable lead over other models like \textit{O4-mini}, \textit{GPT-4.1}, and \textit{DeepSeek-R1}. On the Bamboogle dataset, SOTA LLMs significantly outperform 7B-trained models largely due to the outdated 2018 Wikipedia corpus used in our experiments (see Appendix~\ref{impl_detail} and Figure~\ref{fig:zero_shot_prompt}). %On the Bamboogle dataset, SOTA LLMs significantly outperform 7B-trained models, because Bamboogle requires access to recent Wikipedia content, while our experimental corpus uses the 2018 Wikipedia dump (see Appendix~\ref{impl_detail} and Figure~\ref{fig:zero_shot_prompt}). %SOTA models can compensate for outdated data using their own built-in knowledge.
Overall, these results demonstrate that PVPO consistently achieves state-of-the-art performance across agentic search methods.
% Collectively, these results demonstrate the consistent superiority of PVPO across agentic search methods and affirmatively answer \textbf{Q1} by showing that PVPO achieves state-of-the-art performance compared to existing reinforcement learning techniques.

\subsection{Ablation Study}
\begin{figure}[htbp]
    \centering
    \begin{minipage}[c]{0.42\linewidth}
        \centering
        \setlength{\tabcolsep}{3pt}
        \captionof{table}{Ablation study of PVPO on multi-step retrieval datasets. Starting from DynaSearcher on Qwen2.5-7B-Instruct, we incrementally add Static V Estimation and Group Sampling.
        }
        \label{tab:ablation}
        \begin{tabular}{l cc}
            \toprule
            \multirow{2}{*}{Method} & \multicolumn{2}{c}{\textbf{Average}} \\
            \cmidrule(r){2-3}
            & \textbf{Acc} & \textbf{LasJ} \\
            \midrule
            Qwen2.5-7B-Instruct & 31.1 & 39.7 \\
            GRPO-DynaSearcher & 56.8 & 64.0 \\
            + Static V Estimation & 58.3 & 66.7 \\
            + Group Sampling & 61.0 & 69.6 \\
            \bottomrule
        \end{tabular}
    \end{minipage}
    \hspace{0.03\linewidth}
    \begin{minipage}[c]{0.53\linewidth}
        \centering
        \includegraphics[width=0.85\linewidth]{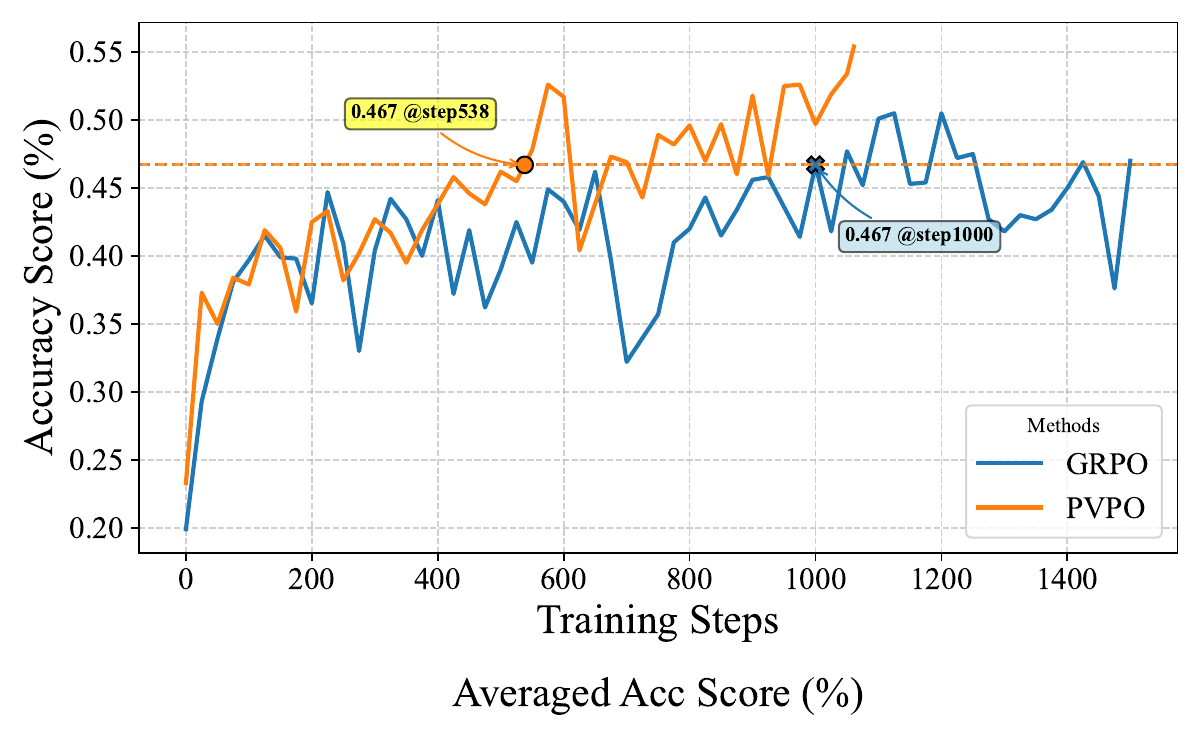}
        \captionof{figure}{Training efficiency of PVPO on mathematical reasoning datasets.}
        \label{fig:EfficiencyAnalysis}
    \end{minipage}
\end{figure}

We conduct an ablation study to isolate the contribution of each component in PVPO, as shown in Table~\ref{tab:ablation}. Starting from the GRPO-DynaSearcher baseline (56.8 Avg Acc / 64.0 LasJ), the integration of Static V Estimation first raises the scores to 58.3/66.7. Subsequently adding Group Sampling further boosts the performance to 61.0/69.6, which represents our full PVPO model and outperforms all baselines. This incremental improvement validates the effectiveness of each proposed component.

% \begin{table*}[tbp] % [htbp] 
%   \centering 
%   \setlength{\tabcolsep}{3pt}
%   \caption{Ablation study of PVPO on multi-step retrieval datasets. Starting from DynaSearcher on Qwen2.5-7B-Instruct, we incrementally add Static V Estimation and Group Sampling.}
%   \label{tab:ablation} 
%   \begin{tabular}{l cc} 
%     \toprule 
%     \multirow{2}{*}{Method} & \multicolumn{2}{c}{\textbf{Avg Acc}} 
%     \\ \cmidrule(r){2-3} & \textbf{Acc} & \textbf{LasJ} \\
%     \midrule
%     Qwen2.5-7B-Instruct      & 31.1 &39.7  \\
%     GRPO-DynaSearcher              & 56.8 &64.0 \\
%     + Static V Estimation          & 58.3 &66.7 \\
%     + Group Sampling                  & 61.0 &69.6 \\
%     \bottomrule 
%   \end{tabular}
% \end{table*}

% ablation  EfficiencyAnalysis

\begin{table*}[tbp] % [htbp] 
  \centering % 
  \setlength{\tabcolsep}{3pt}
  % \caption{Performance comparisons between PVPO and baselines under different scale models on mathematical reasoning datasets, where `w/' represent `with'.}
  \caption{Performance comparison of PVPO and baseline methods on mathematical reasoning datasets using different model scales. ``w/" means trained with.}
  \label{tab:math_performance} % 
  % \begin{tabular}{p{3cm} cccccc} %
  \begin{tabular}{l cccccc} 
    \toprule % 
    
    \textbf{Method} & \textbf{MATH500}  & \textbf{AMC23} & \textbf{Olympiad} & \textbf{AIME-2024} & \textbf{AIME-2025} & \textbf{Avg Acc} \\
    \midrule
    % \multicolumn{7}{l}{\textit{\textbf{Qwen2.5-7B-Instruct}}} \\
    Qwen2.5-7B-Instruct & 75.68	&42.92	&38.94   &12.10	&6.67 &35.26\\
    % w/ GRPO             & 78.66	&49.20	&42.26   &14.26	&12.83 &39.44\\
    w/ GRPO             & 78.60	&49.10	&42.14   &13.86	&10.10 &38.76\\
    w/ DAPO             & 78.58	&51.38	&43.36   &14.96	&11.30 &39.92\\
    % w/ GSPO             & 78.42	&49.98	&43.26   &14.66	&11.54 &39.57\\
    w/ GSPO             & 78.66	&50.12	&43.60   &15.02	&12.70 &40.02\\
    % w/ GSPO             & 79.70	&51.68	&44.20   &12.62	&12.42 &40.12\\ % 869
    % PVPO              & 80.26	&52.06	&44.68  &14.08	&13.50 &40.92\\
    w/ PVPO             & \textbf{80.30}	&\textbf{52.02}	&\textbf{44.62}  &\textbf{14.86}	&\textbf{14.70} &\textbf{41.30}\\
    \midrule
    % \multicolumn{7}{l}{\textit{\textbf{Qwen2.5-14B-Instruct}}} \\
    Qwen2.5-14B-Instruct    &79.68	&51.52	&44.00  &14.82	&12.29  &40.46\\
    % w/ GRPO                &82.22	&55.94	&49.70  &17.14	&16.90  &44.38\\
    w/ GRPO                &82.12	&53.50	&47.42  &16.14	&15.86  &43.01\\
    % w/ DAPO                &82.56	&55.72	&47.52  &16.32	&16.26  &43.68\\ % 210step
    w/ DAPO                &82.50	&56.44	&49.34  &18.04	&15.66  &44.40\\ % 315step
    % w/ GSPO                &82.87	&55.85	&47.95  &16.85	&16.20  &43.94\\
    w/ GSPO                &83.56	&56.02	&49.28  &18.18	&16.20  &44.65\\
    % w/ GSPO                &82.75	&56.55	&49.55  &18.62	&15.88  &44.67\\
    w/ PVPO                &\textbf{83.64}	&\textbf{56.78}	&\textbf{50.72}  &\textbf{19.24}	&\textbf{17.74} &\textbf{45.62}\\
    \bottomrule % 
  \end{tabular}
\end{table*}

\subsection{Generalization Evaluation}
To evaluate the transferability of PVPO, we apply it to mathematical reasoning tasks spanning a range of difficulties, from basic arithmetic to olympiad-level problems. We compare PVPO with GRPO, DAPO, and GSPO across several benchmark datasets.
As shown in Table~\ref{tab:math_performance}, PVPO consistently outperforms all baselines on both the 7B and 14B model scales. % On the 7B model, PVPO surpasses the strongest baseline by at least 1.2 percentage points, and on the 14B model, PVPO leads by at least 1.6 percentage points. PVPO achieves the best results on all five math reasoning datasets, demonstrating clear advantages over the baselines.
% These results confirm that PVPO maintains strong generalization and stable performance across different tasks and domains, directly answering \textbf{Q2}.
We further combine PVPO with the core modules of advanced RL methods, such as the sequence-level importance ratio from GSPO and the KL removal strategy from DAPO, achieving additional performance improvements when integrated with these state-of-the-art algorithms. Since these integrated modules are not the main focus of PVPO, we provide the detailed results and metrics for these extensions in Appendix~\ref{add_exp} and Table~\ref{tab:pvpo_plus}. Furthermore, PVPO exhibits robust cross-domain generalization and enhanced scalability.

% We shifted the task from multi-hop question-answering to single-shot question-answering in the field of mathematical reasoning.
% We selected several mathematical reasoning datasets of varying difficulty, ranging from basic arithmetic reasoning to difficult olympic competition-level problems.
% The aim is to evaluate general reasoning ability in different fields and scenes.
% %
% In the experiment, we compared PVPO with GRPO and the results are shown in Table \ref{tab:math_performance}. 
% First, in terms of average performance, PVPO outperforms at both scales. On the 7B model, PVPO's average score is 1.86 percentage points higher than GRPO's. 
% When the model is scaled up to 14B, PVPO continues to lead.
% % Furthermore, PVPO's advantage is particularly evident on most datasets. For example, on the 7B model, PVPO performance on the AIME-2025 dataset (14.7\%) improved by 14.6\% compared to GRPO (10.83\%).
% The results show that PVPO achieved the best performance on all five mathematical reasoning datasets.
% This fully answers \textbf{Q2}: PVPO has generalizability and maintains stable performance when handling tasks in various fields and scenes.

% 训练步数和验证集指标的图，合并到ablation的表同一行去了
% \begin{figure}[ht]
%     \centering
%     \includegraphics[width=0.8\linewidth]{img/EfficiencyAnalysisFilteredAbnormalv4.pdf}
%     \caption{Training efficiency study of PVPO on mathematical reasoning datasets.}
%     \label{fig:EfficiencyAnalysis}
% \end{figure}

\subsection{Training Efficiency Analysis}
As illustrated in Figure~\ref{fig:EfficiencyAnalysis}, PVPO converges much faster than GRPO, reaching the target accuracy in only 500 steps compared to GRPO's 1,000 steps. After 1,000 steps, PVPO also achieves higher final accuracy, confirming its effectiveness. By applying Group Sampling, PVPO filters out 40--60\% of low-quality data and further accelerates training by 1.7$\times$ to 2.5$\times$ (see Appendix~\ref{filter_ratio}). Overall, these results confirm that PVPO improves both convergence speed and training efficiency.

\begin{figure}[htb]
    \centering
    \includegraphics[width=0.8\linewidth]{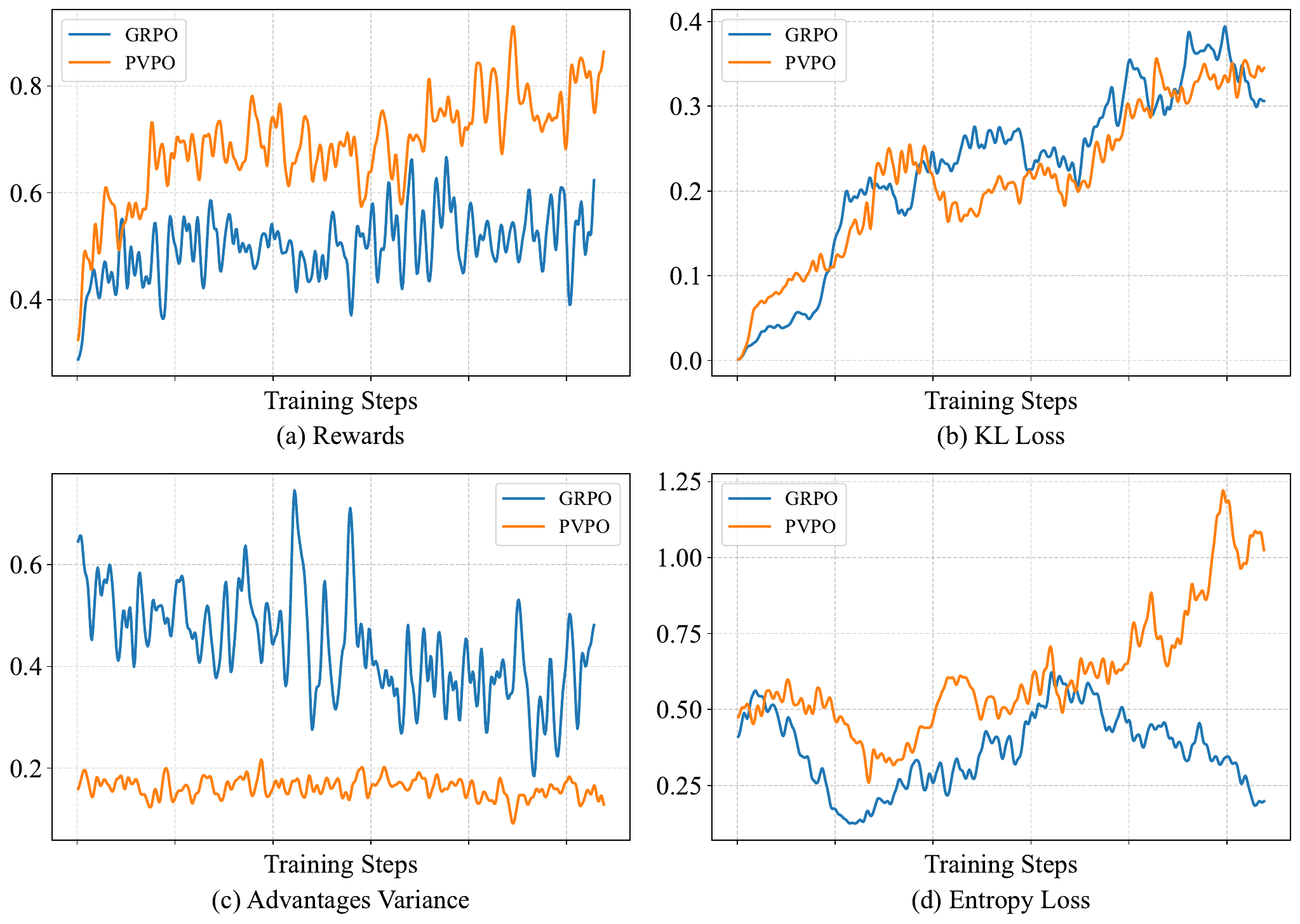}
    \caption{Training stability of PVPO on multi-step retrieval datasets. }
    \label{fig:Stable}
\end{figure}

\subsection{Stability Evaluation}
We track PVPO training metrics to show its stability. Figure~\ref{fig:Stable} (a) shows that PVPO achieves a much higher average reward than GRPO. With a similar KL divergence in Figure~\ref{fig:Stable} (b), this improvement comes not from more aggressive updates, but from better gradient direction estimates.
As shown in Figure~\ref{fig:Stable} (c), PVPO has lower advantage variance, leading to more reliable and consistent update directions.
PVPO also maintains exploration without losing stability. Figure~\ref{fig:Stable} (d) shows that it keeps higher policy entropy under a similar KL constraint, which helps avoid premature convergence to a local optimum.
Overall, PVPO addresses key problems in RL by supporting high exploration, low variance, and high rewards, thereby achieving more stable training than existing methods.
% Overall, PVPO addresses key problems in RL: it supports high exploration, low variance, and high rewards. PVPO achieves more stable training than existing methods.
%This evidence answers \textbf{Q4}: PVPO achieves more stable training than existing methods.

% We systematically tracked PVPO training metrics to demonstrate its stability.
% First, Figure \ref{fig:Stable} (a) shows that PVPO achieves a significantly higher average reward than GRPO. 
% Given the similar KL divergence in \ref{fig:Stable} (b), this gain stems not from more aggressive policy updates, but from higher-quality gradient direction estimates.
% %
% As Figure \ref{fig:Stable} (c) shows PVPO has a significantly lower advantage variance, translating to more consistent and reliable update directions. 
% %
% Moreover, PVPO maintains exploration without sacrificing stability. As shown in Figure \ref{fig:Stable} (d), it maintains higher policy entropy under a similar KL constraint, effectively preventing premature convergence to a local optimum.
% Overall, PVPO truly solves the core problems of high exploration, low variance, and high rewards in RL.
% This evidence collectively answers \textbf{Q4}: PVPO can achieve more stable training dynamics than existing methods.

\subsection{Case Study: Low Sampling Budget}
\begin{figure}[t]
    \centering
    \includegraphics[width=0.85\linewidth]{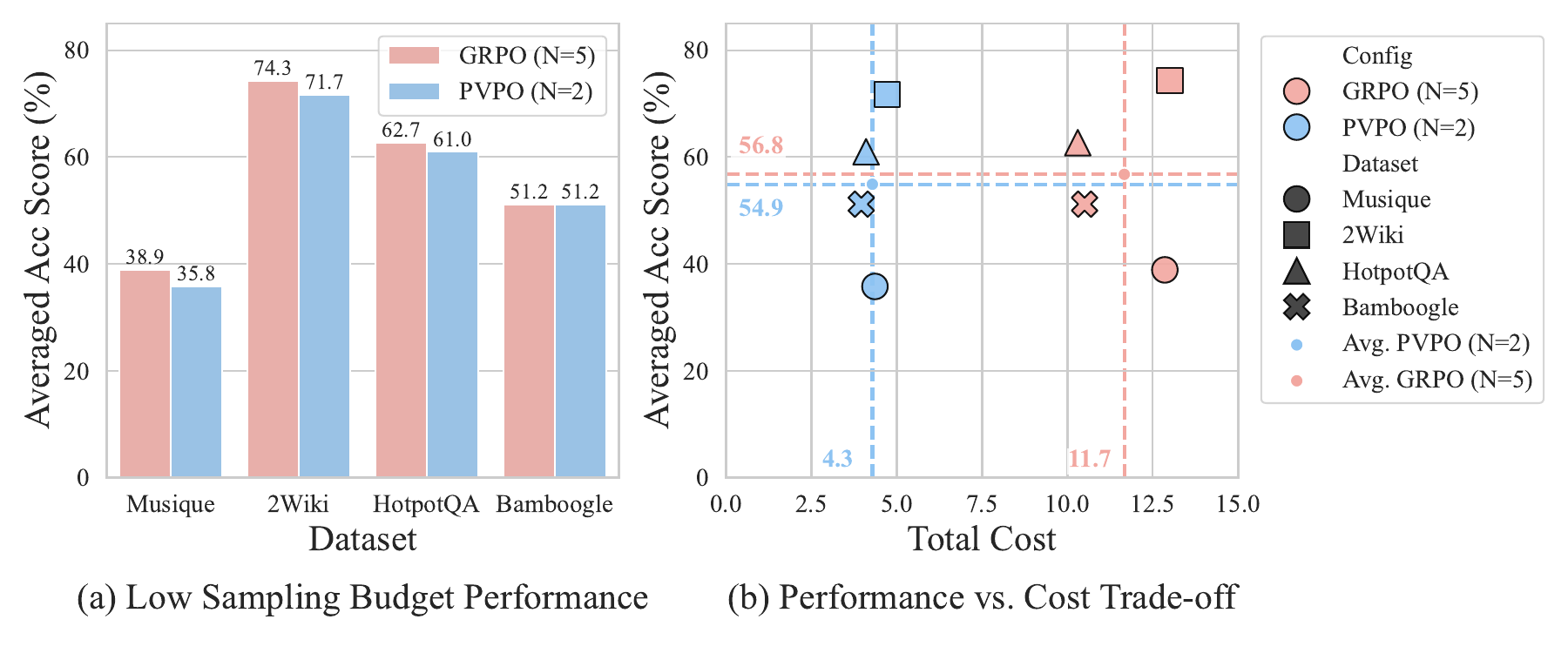}
    \caption{Low sampling budget of PVPO on multi-step retrieval datasets. The $N$ denotes the number of trajectories in each single rollout. $N$=5 is the full budget and $N$=2 is the low budget.}
    \label{fig:CaseStudy}
\end{figure}

% We ran a case study to address \textbf{Q5}. 
To further examine PVPO's performance under resource constraints, we conduct a case study on low sampling budget.
We reduce the number of rollouts from 5 (used in the main experiments) to 2. For comparison, we report GRPO's performance with a full budget.
Figure~\ref{fig:CaseStudy} (a) shows that PVPO with a low budget remains close to the fully budgeted GRPO. We calculate computational cost by multiplying the number of rollouts with the average number of tool calls in trajectories. As shown in Figure~\ref{fig:CaseStudy} (b), PVPO's average cost is only 4.3, which is much lower than GRPO's 11.7.
PVPO achieves 97\% of GRPO's performance (55.0\% vs 56.8\%) while using less than 40\% of the computational cost. This strong sample efficiency comes from the high-quality, low-variance training signals provided by Static V Estimate. The model can update its policy efficiently using fewer rollouts.
% This answers \textbf{Q5}: model training clearly benefits from Static V Estimate.

% We conducted a case study to explore \textbf{Q5}.
% We reduced the number of rollouts from 5 (as used in our main experiments) to 2.
% %We then re-evaluated PVPO's performance on the multi-hop question-answering task. 
% For comparison, we report the performance of GRPO in a full budget setting.
% %The results presented in Figure \ref{fig:CaseStudy}.
% As shown in Figure \ref{fig:CaseStudy} (a), PVPO in low budget setting remains close to the fully budgeted GRPO.
% Our cost calculation method is based on the number of rollouts multiplied by the average number of tool calls in trajectories.
% As shown in Figure \ref{fig:CaseStudy} (b), the average computational cost of PVPO is only 4.3, significantly lower than GRPO's 11.7. 
% PVPO achieved 97\% of GRPO's performance (55.0\% vs. 56.8\%) using less than 40\% of computational cost. 
% %解释机制
% This outstanding sample efficiency stems from the high-quality, low-variance training signals provided by our Static V Estimate.
% Consequently, the model can efficiently update its policy with only a few rollouts.
% This answers \textbf{Q5}: model training significantly benefit from Static V Estimate.

\section{Conclusions}
In this paper, we propose PVPO, an efficient critic-free reinforcement learning algorithm designed to optimize policy learning for complex tasks. By introducing a Static V Estimate as an external advantage reference and integrating it with group sampling for effective data filtering, PVPO addresses the limitations of extensive sampling and biased intra-group comparisons inherent in prior methods. Our approach yields stable, low-variance training signals, accelerates convergence, and significantly reduces computational costs.
% Our approach leads to stable, low-variance training signals, faster convergence, and much lower computational cost. 
Extensive experiments across nine diverse benchmarks in multi-hop QA and mathematical reasoning demonstrate that PVPO achieves state-of-the-art performance and strong generalization, even with small-scale models and limited resources. 
% PVPO offers reliable reasoning and tool-use improvements, scalable training, and consistent performance, demonstrating its substantial potential for broader real-world applications.
PVPO introduces substantial improvements in reasoning and tool use, supports scalable training, and ensures consistent performance, thereby demonstrating strong potential for widespread real-world application.

% In this paper, we propose PVPO, a new RL algorithm that redefines advantage estimation to cut down computational cost and gradient variance. Static V Estimate provides stable policy guidance by decoupling the advantage baseline from sampling. Group Sampling shortens training time by pruning the dataset efficiently.
% Extensive experiments on nine complex datasets from two domains show PVPO's effectiveness and strength. As a critic-free method, PVPO is fast to train and generalizes well. It also reduces the need for massive sampling in on-policy methods.

% This paper proposes a novel RL algorithm PVPO, which redefines the advantage estimation to significantly reduce computational cost and gradient variance. 
% Static V Estimate provides stable policy guidance by decoupling the advantage baseline from the sampling process. Group Sampling strategy shortens training time through efficient dataset pruning.
% Extensive experiments conducted on nine complex datasets across two domains fully validate the superiority and effectiveness of PVPO. As a method that follow the critic-free paradigm, PVPO features fast training speed and strong generalization ability. It effectively mitigates the reliance of current on-policy methods on massive sampling.

%\subsubsection*{Acknowledgments}
%All acknowledgments, including those to funding agencies, go at the end of the paper.

\bibliography{iclr2026_conference}
\bibliographystyle{iclr2026_conference}

\clearpage % 强制开启一个新页面
\appendix
\section{Appendix}
\subsection{Implementation Details}\label{impl_detail}
\textbf{Retriever and Corpus}. For the multi-hop QA task, we employ \textit{multilingual-e5-base} as the retriever model and use the December 2018 Wikipedia dump as the primary retrieval corpus, which contains over 21 million passages. To improve retrieval efficiency, we construct the final corpus by combining supporting document passages from three multi-hop datasets (i.e., Musique, 2Wiki, and HotpotQA) with one million randomly sampled documents from the Wikipedia dump. Notably, Bamboogle only provides questions and answers without ground truth passages, so it cannot be incorporated into the retrieval corpus. This may contribute to the lower scores on Bamboogle for most methods, as shown in Table~\ref{tab:QA_performance}.%, since SOTA LLMs may have seen related knowledge during pre-training or post-training. 
Passage retrieval is implemented using FAISS\footnote{\url{https://pypi.org/project/faiss-gpu/}}, and for each query, the top 5 passages are retrieved during both training and testing. For the KG (Knowledge Graph) data used in \textbf{PVPO-DynaSearcher}, we follow the approach and dataset provided by \citet{wang2021kepler}, which is aligned with \citet{hao2025dynasearcherdynamicknowledgegraph}.

\textbf{Prompts and Code}. We implement \textbf{PVPO-ReSearch} and \textbf{PVPO-DynaSearcher} based on the ReSearch framework\footnote{\url{https://github.com/Agent-RL/ReCall/tree/re-search}}. The system prompts for ReSearch and DynaSearcher are set following their respective original papers, detailed prompt templates are shown in Figure~\ref{fig:ReSearch_prompt} and ~\ref{fig:dynasearcher_prompt}. For prompt-based SOTA LLMs, we first retrieve 5 passages from the corpus for each question, and then organize these passages using the template shown in Figure~\ref{fig:zero_shot_prompt} as the prompt for answer generation.
For mathematical reasoning tasks, we use \texttt{verl} version \texttt{0.3.1.dev0}. 
Since the ReSearch codebase is also developed on top of the verl framework, we provide the core implementation of our PVPO method based on verl in code Listing~\ref{list:pvpo_code}.
%The core implementation of our PVPO method based on the verl framework is provided in Listing~\ref{lst:pvpo_code}.

% \iffalse
\begin{center}
\begin{minipage}{0.93\textwidth}
% \centering
\begin{lstlisting}[language=Python, caption={PyTorch-style pseudocode for PVPO}, label={list:pvpo_code}]
# verl/trainer/ppo/ray_trainer.py
def compute_advantage(...):
    if adv_estimator == AdvantageEstimator.PVPO:
        # compute pvpo advantages
        advantages, returns = core_algos.compute_pvpo_outcome_advantage(
            token_level_rewards=data.batch["token_level_rewards"],
            token_level_values=data.non_tensor_batch["static_value"],
            response_mask=data.batch["response_mask"],
        )
        data.batch["advantages"] = advantages
        data.batch["returns"] = returns
    ...
# verl/trainer/ppo/core_algos.py
def compute_pvpo_outcome_advantage(
    token_level_rewards: torch.Tensor,
    token_level_values: torch.Tensor,
    response_mask: torch.Tensor,
):
    scores = token_level_rewards.sum(dim=-1)
    values = torch.tensor(token_level_values.astype(np.float32), device=scores.device, dtype=scores.dtype)

    with torch.no_grad():
        for i in range(scores.shape[0]):
            scores[i] = (scores[i] - values[i])
        scores = scores.unsqueeze(-1) * response_mask
    return scores
\end{lstlisting}
\end{minipage}
\end{center}
% \fi
% \begin{algorithm}[H]
% \caption{PVPO Advantage Calculation}
% \begin{algorithmic}[1]
% \Function{ComputeAdvantage}{data, adv\_estimator}
%     \If{adv\_estimator == PVPO}
%         \State $(\text{advantages}, \text{returns}) \gets$ \Call{ComputePVPOOutcomeAdvantage}{
%             data.batch["token\_level\_rewards"], \\
%             data.non\_tensor\_batch["static\_value"], \\
%             data.batch["response\_mask"]}
%         \State data.batch["advantages"] $\gets$ advantages
%         \State data.batch["returns"] $\gets$ returns
%     \EndIf
% \EndFunction
% \vspace{1ex}
% \Function{ComputePVPOOutcomeAdvantage}{rewards, values, mask}
%     \State scores $\gets$ Sum(rewards, dim=$-1$)
%     \State values $\gets$ ConvertToTensor(values)
%     \For{each $i$ in batch}
%         \State scores[$i$] $\gets$ scores[$i$] $-$ values[$i$]
%     \EndFor
%     \State scores $\gets$ Unsqueeze(scores, dim=$-1$) $\times$ mask
%     \State \Return $(\text{scores}, \text{scores})$
% \EndFunction
% \end{algorithmic}
% \end{algorithm}

% \subsection{Prompt Examples for Multi-hop QA}

\begin{figure*}[htbp]
\begin{tcolorbox}[
    colback=gray!20,  % 20% gray,
    colframe=black,
    ] % title=
You are an expert in question answering. Given a question within \texttt{<question>} \texttt{</question>} \
and some contexts within \texttt{<context>} \texttt{</context>}, you first think about the reasoning process within \texttt{<think>} \texttt{</think>} \
and put the answer within \texttt{<answer>} \texttt{</answer>}. For example, <question> This is a question \texttt{<question>} \
\texttt{<context>} Here are contexts \texttt{<context>} \texttt{<think>} This is the reasoning process. \texttt{</think>} \
\texttt{<answer>} The final answer is \text{\textbackslash boxed\{ answer here \}} \texttt{</answer>}. \
If the answer could not be deduced from the contexts or it's wrong, give the right answer based on your own knowledge. \
In the last part of the answer, the final exact answer is enclosed within \verb|\boxed{}|.
\end{tcolorbox}
\caption{Prompt for zero-shot LLM RAG.}
\label{fig:zero_shot_prompt}
\end{figure*}

\begin{figure*}[htbp]
\begin{tcolorbox}[
    colback=gray!20,  % 20% gray,
    colframe=black,
    ] % title=
You are a helpful assistant that can solve the given question step by step with the help of the wikipedia search tool. \
Given a question, you need to first think about the reasoning process in the mind and then provide the answer. \
During thinking, you can invoke the wikipedia search tool to search for fact information about specific topics if needed. \
The reasoning process and answer are enclosed within \texttt{<think>} \texttt{</think>} and \texttt{<answer>} \texttt{</answer>} tags respectively, \
and the search query and result are enclosed within \texttt{<search>} \texttt{</search>} and \texttt{<result> </result>} tags respectively. \
For example, \texttt{<think>} This is the reasoning process. \texttt{</think>} \texttt{<search>} search query here \texttt{</search>} \texttt{<result>} search result here \texttt{</result>} \
\texttt{<think>} This is the reasoning process. \texttt{</think>} \texttt{<answer>} The final answer is \text{\textbackslash boxed\{ answer here \}} \texttt{</answer>}. \
In the last part of the answer, the final exact answer is enclosed within \verb|\boxed{}|.
\end{tcolorbox}
\caption{System prompt for ReSearch.}
\label{fig:ReSearch_prompt}
\end{figure*}

\begin{figure*}[htbp]
\begin{tcolorbox}[
    colback=gray!20,  % 20% gray,
    colframe=black,
    ] % title=
You are a helpful assistant that can solve the given question step by step with the help of the wikipedia search tool. \
Given a question, you need to first think about the reasoning process in the mind and then provide the answer. \
During thinking, you can invoke the wikipedia search tool to search for fact information about specific topics if needed. \
The reasoning process and answer are enclosed within \texttt{<think>} \texttt{</think>} and \texttt{<answer>} \texttt{</answer>} tags respectively, \
and the search input and result are enclosed within \texttt{<search>} \texttt{</search>} and \texttt{<result> </result>} tags respectively. \
Search input is json format like \{``query'': ``xxx'', ``entity'': [``yyy''], ``relation'': [``zzz'']\} and applied to the search tools, \
where query is used to search wikipedia articles, entity(s) and relation(s) are used to search wikidata, a knowledge base of entities and relations. \

For example, \texttt{<think>} This is the reasoning process. \texttt{</think>} \texttt{<search>} \{``query'': ``Who is the director of Avatar'', ``entity'': [``Avatar''], ``relation'': [``director'']\} \texttt{</search>} \texttt{<result>} search result here \texttt{</result>} \
\texttt{<think>} This is the reasoning process. \texttt{</think>} \texttt{<answer>} The final answer is \text{\textbackslash boxed\{ answer here \}}\texttt{</answer>}. \
In the last part of the answer, the final exact answer is enclosed within \verb|\boxed{}|. 
\end{tcolorbox}
\caption{System prompt for DynaSearcher.}
\label{fig:dynasearcher_prompt}
\end{figure*}

\begin{figure*}[ht]
\begin{tcolorbox}[
    colback=gray!20,  % 20% gray,
    colframe=black,
    ] % title=Prompt for LasJ.
You will be provided with three pieces of content: the questioner's question, the user's response, and the reference answer list.
Your task is to score the accuracy of the user's response based on the criteria outlined below.
Please ensure that you carefully read and understand these instructions.
Evaluation Criteria:
1. The pred answer doesn’t need to be exactly the same as any of the ground truth answers, but should be semantically same for the question.
2. Each item in the ground truth answer list can be viewed as a ground truth answer for the question, and the pred answer should be semantically same to at least one of them.
3. The user's response may be longer and more detailed; as long as it is logically correct, contains the correct answer, it should be scored appropriately.
Evaluation Steps:
1. Carefully read the questioner's question and understand its key points.
2. Carefully read the reference answer and understand the key points relevant to the question.
3. Based on the evaluation criteria, assign a score in the range of 0 to 5, where 0 indicates that the user's response does not include any of \
the key points from the reference answer and completely fails to answer the questioner's question; 5 indicates that the user's response includes \
all the key points from the reference answer and fully and correctly answers the questioner's question.

Questioner's question:
\{question\}\\
Reference answer:
\{answer\}\\
User's response:
\{response\}

Evaluation result (output only the score between 0 and 5):
\end{tcolorbox}
\caption{Prompt for LLM-as-Judge score.}
\label{fig:lasj_prompt}
\end{figure*}

\subsection{Group Sampling Analysis}\label{filter_ratio}
\begin{figure}[t]
    \centering
    \includegraphics[width=0.8\linewidth]{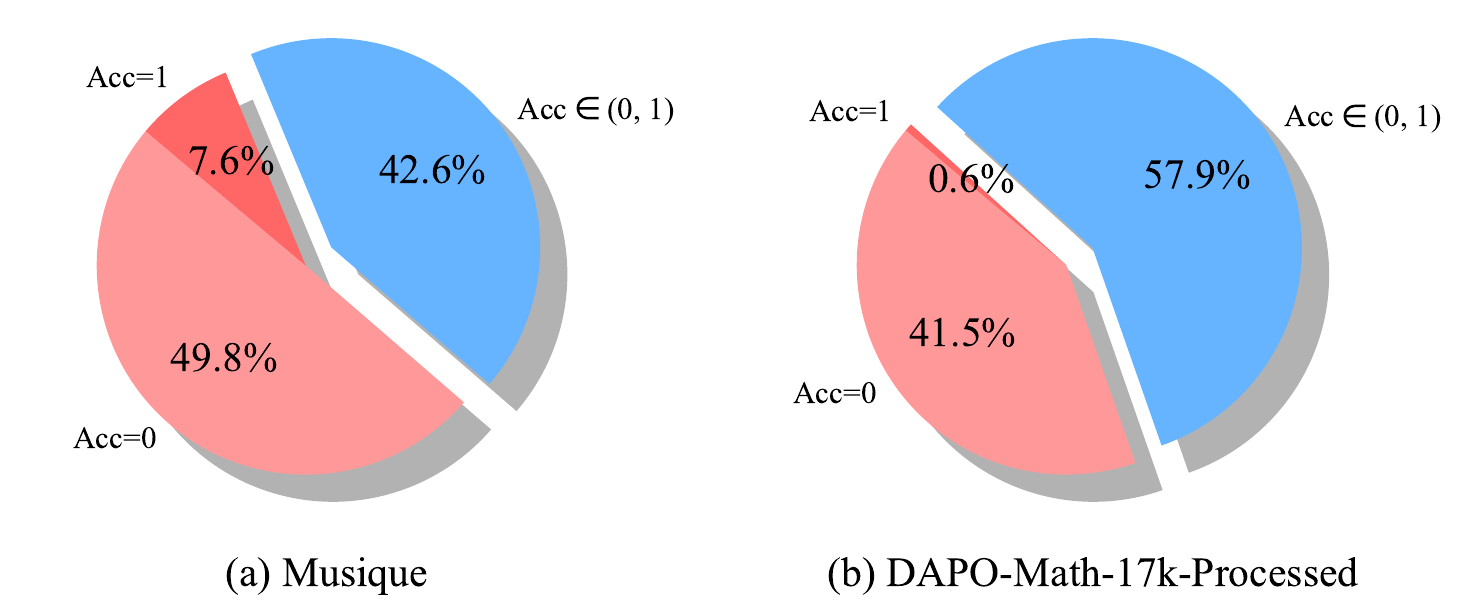}
    \caption{Group Sampling study on datasets from different fields. The Acc is the mean of the answer accuracies from M trajectories rolled out by the reference model. 
    M=5 in Figure (a) and M=16 in Figure (b).}
    \label{fig:FilterData}
\end{figure}

We calculate the data filtering ratio on two training sets, as shown in Figure~\ref{fig:FilterData}. Group Sampling removes samples with Acc = 1 or 0 before training, filtering out 40\%-60\% of the total dataset. This leads to a 1.7--2.5$\times$ increase in training efficiency.
% This does not affect performance, since the model gets effective gradients each iteration after low-value samples are removed. Also, Group Sampling only runs inference once, so the extra overhead is much less than the training cost of keeping low-value samples.
% Experiments show that PVPO greatly reduces total training time.
% In summary, this answers \textbf{Q3}: PVPO can reduce the temporal consumption of training resources.

\subsection{Additional Experiment Results}\label{add_exp}
To further verify the scalability of our proposed PVPO method,  we conduct integration experiments on multi-hop QA tasks. Specifically, we combine PVPO with the sequence-level importance ratio module proposed in GSPO and remove the KL loss constraint as introduced in DAPO.
The results, shown in Table~\ref{tab:pvpo_plus}, demonstrate that PVPO not only provides strong baseline improvements over GRPO, but also achieves further performance gains when integrated with these advanced RL methods. In particular, the combination with DAPO (w/o KL) yields the best accuracy and LasJ scores, while integration with GSPO’s sequence-level importance ratio also presents consistent improvements. In particular, the combination with DAPO (w/o KL) yields the best accuracy and LasJ scores, but also incurs significantly more tool calls (8.14 per query), resulting in greater inference costs. By contrast, GSPO’s sequence-level importance ratio offers improvements with relatively lower tool call overhead (2.19 per query). Therefore, the trade-off between performance and inference cost should be considered when choosing an integration strategy for different practical scenarios.
These findings confirm that PVPO is highly compatible and complementary when used alongside other state-of-the-art RL algorithms.

% \clearpage
\begin{table*}[htbp] % [htbp] 
  \centering 
  \setlength{\tabcolsep}{3pt}
  \caption{Experimental results of PVPO's orthogonal integration with SOTA RL algorithms (DAPO, GSPO) and scalability evaluation on multi-hop QA tasks. ``w/ Seq-Ratio" refers to the sequence-level importance ratio from GSPO, and ``w/o KL" means removing the KL loss constraint as in DAPO.}
  \label{tab:pvpo_plus} 
  \begin{tabular}{l ccc} 
    \toprule 
    \multirow{2}{*}{Method} & \multicolumn{3}{c}{\textbf{Average}} 
    \\ \cmidrule(r){2-4} & \textbf{Acc} & \textbf{LasJ} & \textbf{ToolCalls} \\
    % Method & \textbf{Acc} & \textbf{LasJ} & \textbf{ToolCalls} \\
    \midrule
    GRPO-ReSearch         & 48.6 &58.0 & 2.46 \\
    PVPO-ReSearch         & 54.4 &62.6 & 2.96 \\
    w/ Seq-Ratio (GSPO)   & 55.1 &62.4 & 2.19 \\
    w/o KL (DAPO)         & \textbf{58.8} &\textbf{67.1} & 8.14 \\
    \bottomrule 
  \end{tabular}
\end{table*}

\end{document}